\documentclass{article}

\usepackage[english]{babel}

\usepackage[letterpaper,top=2cm,bottom=2cm,left=3cm,right=3cm,marginparwidth=1.75cm]{geometry}

\usepackage{amsmath}
\usepackage{graphicx}
\usepackage[colorlinks=true, allcolors=blue]{hyperref}

\usepackage{multirow}
\usepackage{caption}
\usepackage{subcaption}
\usepackage{float}

\newcommand{\bx}{\mathbf{x}}

\newcommand{\bw}{\mathbf{w}}

\title{Solving Partial Differential Equations with Point Source Based on Physics-Informed Neural Networks}
\author{Xiang Huang$^{1}$,
	Hongsheng Liu$^{2}$,  Beiji Shi$^{2}$,  Zidong Wang$^{2}$,  Kang Yang$^{2}$, \\ Yang Li$^{2}$,  Bingya Weng$^{2}$,  Min Wang$^{2}$, Haotian Chu$^{2}$, Jing Zhou$^{2}$,  Fan Yu$^{2}$, \\ Bei Hua$^{1}$,  Lei Chen$^{3}\thanks{Corresponding author.}$,  Bin Dong$^{4}$\\ 
	\small{$^1$University of Science and Technology of China},
	\small{$^2$Huawei Technologies Co. Ltd}\\
	\small{$^3$Hong Kong University of Science and Technology},
	\small{$^4$Peking University}\\
	\small{$^1$sahx@mail.ustc.edu.cn, bhua@ustc.edu.cn}\\
	\small{$^2$liuhongsheng4, shibeiji, wang1, yangkang22, liyang477@huawei.com}\\
		\small{$^2$wengbingya, wangmin106, chuhaotian2, zhoujing3, fan.yu@huawei.com}\\
		\small{$^3$leichen@cse.ust.hk},	\small{$^4$dongbin@math.pku.edu.cn}
	
}

\date{}

\begin{document}

\maketitle

\begin{abstract}
In recent years, deep learning technology has been used to solve partial differential equations (PDEs), among which the physics-informed neural networks (PINNs) emerges to be a promising method for solving both forward and inverse PDE problems. 
PDEs with a point source that is expressed as a Dirac delta function in the governing equations are mathematical models of many physical processes. 
However, they cannot be solved directly by conventional PINNs method due to the singularity brought by the Dirac delta function.
We propose a universal solution to tackle this problem with three novel techniques. 
Firstly the Dirac delta function is modeled as a continuous probability density function to eliminate the singularity; 
secondly a lower bound constrained uncertainty weighting algorithm is proposed to balance the PINNs losses between point source area and other areas; 
and thirdly a multi-scale deep neural network with periodic activation function is used to improve the accuracy and convergence speed of the PINNs method. 
We evaluate the proposed method with three representative PDEs, and the experimental results show that our method outperforms existing deep learning-based methods with respect to the accuracy, the efficiency and the versatility.
\end{abstract}

\section{Introduction}

\noindent Partial differential equations (PDEs) play a significant role in science and engineering disciplines, which are used to formulate physical problems such as sound propagation, fluid flow, electromagnetic field, etc. 
In the past decades, numerical solutions of PDEs have been developed and successfully applied to many real-world applications like aerodynamics and weather forecasting.
These numerical methods, two representatives of which are finite difference method (FDM) and finite element method (FEM), firstly mesh the solution areas of the problem, and then obtain the approximate values of the exact solution on the discrete grid points. 
As the scale and complexity of the problem increase, the computational complexity of numerical methods becomes prohibitively high. 

On the other hand, with the increase use of numerical methods in both industry and academia, a growth in the collection of simulation data has been observed in computational fluid dynamics, electromagnetic simulation and many other areas \cite{ansari2018data,yu2019integrated}, and data-driven deep learning approaches are proposed to alleviate the computational burden. 
For example, Thuerey et al. \cite{thuerey2020deep} used U-Net to predict the velocity and pressure fields around an airfoil. 
Khan et al. \cite{khan2019deep} trained a convolutional neural network to solve Maxwell's equations for low-frequency electromagnetic devices.
Lu et al. \cite{lu2019deeponet} proposed a deep operator network (DeepONet) that can learn any mathematical operator (mapping a function to another function) based on the universal approximation theorem of operators.
For PDEs, neural operators directly learn the mapping from any functional parametric dependence to the solution. 
Li et al. \cite{li2020fourier} proposed the Fourier neural operator (FNO) that parameterizes the integral kernel directly in Fourier space, allowing for an expressive and efficient architecture. 
FNO was applied to solve Burgers' equation, Darcy flow, and Navier-Stokes equation. 

Data-driven deep learning methods have the advantage that once the neural network is trained, the prediction process is fast.
However, the cost of data acquisition is prohibitive for complex physical, biological, and engineering systems, and the generalization ability of the model (from one experimental condition to another) is poor when there are not sufficient labeled data \cite{cai2021physics}.
Therefore, physics-informed solutions to PDEs have been proposed in recent years, which rely on the universal approximation ability of deep neural network \cite{pinkus1999approximation} and leverage the advantage of the automatic differentiation in the deep learning frameworks \cite{baydin2018automatic}. 
Compared with traditional PDE solvers, deep learning solvers \cite{lagaris1998artificial,sirignano2018dgm,weinan2018deep,raissi2019physics} not only can deal with parametrized simulations, but also address the inverse or data assimilation problems that the traditional solvers cannot deal with. 
Among these methods, the physics-informed neural networks (PINNs) \cite{raissi2019physics} attracts increasing  attentions from the machine learning and applied mathematics communities due to its simple and effective way of approximating the solutions of the PDEs using neural networks. 
The PINNs method preserves the physical information in the governing equations and adapts to the boundary conditions. 
The PINNs method has been applied to solve many well-known PDEs such as  Navier-Stokes equations \cite{jin2021nsfnets}, Schroedinger's equation \cite{raissi2018deep}, and Maxwell's equations \cite{zhang2021maxwell}.

PDEs with a point source that is expressed mathematically as a Dirac delta function $\delta(\bx)$ have many applications in physical simulation.
For instance, a point source can be a pulse exciting the electric field in electromagnetic simulation \cite{sullivan2013electromagnetic}, or a sound source in an acoustic wave equation \cite{moseley2020solving}.
When we use the residuals of PDEs with the point source as the loss term (i.e., the direct application of the conventional PINNs method), the training will not converge due to the singularity brought by the Dirac delta function.
Several methods have been proposed in the literature to solve this kind of problem under some specific conditions.
For example, the Deep Ritz method \cite{weinan2018deep} and the variational method in NVIDIA SimNet \cite{hennigh2021nvidia} need to transform the point source term into a computer computable form, which however is not applicable to all PDEs with a point source;
PINNs-based methods proposed by Zhang et al. \cite{zhang2021maxwell}, Moseley et al. \cite{moseley2020solving}, and Bekele tet al. \cite{bekele2020physics} require a significant amount of labeled data that may not be available in some cases.

In this paper, we propose a universal solution to the point source PDE problem based on PINNs method. 
The proposed method does not rely on any labeled data or variational forms, and outperforms existing deep learning-based methods with respect to the accuracy, the efficiency and the versatility. 
The main features of our method are in three aspects:

\begin{itemize}
	\item The Dirac delta function is modeled as a continuous probability density function to eliminate the singularity;
	\item A lower bound constrained uncertainty weighting algorithm is proposed to balance the loss term between the point source area and other areas;
	\item A neural network architecture is built with multi-scale DNN \cite{xu2019frequency} and periodic activation functions to improve the accuracy and convergence speed of the PINNs method.
\end{itemize}

The rest of this paper is organized as follows: 
Sec.\ref{sec:preliminaries} briefly introduces relevant preliminary knowledge and related works, 
Sec.\ref{sec:methodology} describes our method under the PINNs framework, 
Sec.\ref{sec:numerical_experiments} performs extensive experiments on the method with various benchmark problems, 
and Sec.\ref{sec:conclusions} concludes the paper.

\section{Preliminaries}\label{sec:preliminaries}
The general form of a time-dependent PDE is given by :
\begin{equation}\label{def:point_src_pde}
	\begin{aligned}
		\mathcal{N}(u(\bx, t)) &=f(\bx, t),  \quad (\bx, t) \in \Omega\times [0, T] \\
		u(\bx, t) &= g(\bx, t),  \quad (\bx, t) \in \partial\Omega\times [0, T]  \\
		u(\bx, 0) & = h(x),  \qquad \bx \in \Omega
	\end{aligned}
\end{equation}
where  $\mathcal{N}(\cdot)$ is the differential operator, $u$ is the solution, and $f(\bx, t)$ is the time-dependent forcing term. 
For the PDEs with point source we consider in this paper, the forcing term $f(\bx,t)$ contains the Dirac delta functions.

\subsection{PDEs with Point Source}
Here we introduce two well-known time-(in)dependent PDEs with point source forcing terms.
The Poisson's equation with a point source has the following form:
\begin{equation}\label{def:possion_point_source}
	\begin{aligned}
		-\Delta u &=\delta_{}(\bx-\bx_0),  \quad \bx \in \Omega, \\
		u&= 0,  \qquad\qquad\,\, \,\, \bx \in \partial\Omega
	\end{aligned}
\end{equation}
where $\delta$ denotes the Dirac delta function. 
Although it looks simple, Poisson's equation is a fundamental model for describing diffusion process, such as fluid flow, heat transfer, and chemical transport.
In the field of electromagnetic simulations, the time domain Maxwell's equations with excitation can be expressed as 
\begin{equation}\label{def:maxwell_point_source}
	\begin{aligned}
		\nabla\times E&=-\mu \dfrac{\partial H}{\partial t} - J(\bx, t), \\
		\nabla\times H&=\epsilon \dfrac{\partial E}{\partial t}.
	\end{aligned}
\end{equation}
where $\epsilon$ is the permittivity, $\mu$ is the permeability. The forcing term $J(\bx, t)=\delta(\bx - \bx_0)g(t)$ excites a wave propagation at $\bx_0$ with a particular type of pulse $g(t)$.  
Maxwell's equations survived the revolutionary changes in physics in the 20th century and they underpin a huge range of natural phenomena.

\subsection{The PINNs Method}
In order to solve PDE problems with deep learning technology
the PINNs method is proposed to approximate the solution $u(\bx, t)$ with a deep neural network $u(\bx, t;\theta)$, where $\theta$ represents the trainable parameters of the deep neural network. 
The loss function given by such approximation is defined as 
\begin{subequations}\label{eq:PINN_loss}
\begin{align}
	L_{total}(\theta) &= L_{r}(\theta) + \lambda_{ic}L_{ic}(\theta)+\lambda_{bc}L_{bc}(\theta),\label{eq:PINN_loss_total}\\
	L_{r}(\theta) &= \dfrac{1}{N_{r}}\sum_{i=1}^{N_{r}} ||\mathcal{N}(u(\bx_i, t_i;\theta)) -f(\bx_i, t_i)||_2^2,\label{eq:PINN_loss_pde}\\
	L_{ic}(\theta) &= \dfrac{1}{N_{ic}}\sum_{i=1}^{N_{ic}} ||u(\bx_i, 0;\theta)  -h(\bx_i)||_2^2, \label{eq:PINN_loss_ic}\\
	L_{bc}(\theta)& =\dfrac{1}{N_{bc}}\sum_{i=1}^{N_{bc}} ||u(\bx_i, t_i;\theta)  -g(\bx_i, t_i)||_2^2. \label{eq:PINN_loss_bc}
\end{align}
\end{subequations}
where $L_{r}$, $L_{ic}$, $L_{bc}$ correspond to the loss term of PDE residual, initial conditions, and boundary conditions, respectively.
The hyperparameters $\lambda_{ic}$ and $\lambda_{bc}$ aim to balance the interplay of different terms in the loss function. 
$N_r$, $N_{ic}$, and $N_{bc}$ represent the number of samples in the entire region, initial conditions, and boundary conditions, respectively.

When $f(\cdot), g(\cdot), h(\cdot)$ are continuous, the network weights $\theta$ can be optimized by minimizing the total training loss $L_{total}(\theta)$ via standard gradient descent procedures used in deep learning.  
Unfortunately, due to the singularity brought by $\delta(\bx)$, the PINNs method cannot be directly used to solve the PDEs with a point source.

\subsection{Related Works}\label{subsec:related_works}
One can use the Deep Ritz method \cite{weinan2018deep} to solve Eq.\eqref{def:possion_point_source}.
Through mathematical derivation, Eq.\eqref{def:possion_point_source} can be converted to the minimum value problem:
\begin{equation}\label{def:possion_point_source_ritz}
	\begin{aligned}
       \min \int_{\Omega}{\frac{1}{2}\| \Delta u(\bx)\|^2 d\bx - u(\bx_0)} + \int_{\partial \Omega}{\|u(\bx)\|^2}d\bx
	\end{aligned}
\end{equation}
where the Dirac function term $\delta(\bx-\bx_0)$ in Eq.\eqref{def:possion_point_source} will generate the constant term $u(\bx_0)$ in Eq.\eqref{def:possion_point_source_ritz} after integration. 
However, application of the Deep Ritz method is restricted because it's unable to handle PDEs like Eq.\eqref{def:maxwell_point_source} which cannot be derived from any variational problems.
The NVIDIA SimNet \cite{hennigh2021nvidia} proposes to solve PDEs with a point source using the variational formulation.
However, the performance heavily depends on the selection of the test functions and the computational time increases linearly with the number of test functions. 

Zhang et al. \cite{zhang2021maxwell} and Moseley et al. \cite{moseley2020solving} use the classical numerical method to simulate the early period ($0\leq t \leq T_0 < T$) of wave equations to avoid the point source singularity, and then applied the PINNs method to solve the PDEs with given initial conditions at $t=T_0$.
Bekele et al. \cite{bekele2020physics} attempted to leverage the PINNs loss by adding  a new loss term with a significant amount of known labels when solving the  Barry and Mercer's source problem with time-dependent fluid injection.  
However, this method relies on classical computational methods to generate the labeled data which is expensive or even infeasible in many situations.

In this work, we propose a universal solution based on the PINNs method which does not rely on any labeled data or variational forms. 
Furthermore, our method greatly improves the accuracy and convergency speed compared with current deep learning based approaches.
\section{Methodology}\label{sec:methodology}
In this section, we explain our strategy to tackle the singularity problem in detail.

\subsection{Approximation to $\delta(\bx)$}\label{subsec:Smooth}
The Dirac delta function $\delta(x)$ is mathematically defined by: 
\begin{equation}\label{def:delta_function}
\delta(x) =\begin{cases}
+\infty,  & x = 0 \\
0,  & x \neq 0
\end{cases}
\end{equation}
and which is also constrained to satisfy the identity $\int_{-\infty}^{+\infty}\delta(x)dx = 1$.
Taking $\eta(x)$ to be any symmetric unimodal continuous probability density function centered at 0, $\eta(x)=\alpha^{-1}\eta(\dfrac{x}{\alpha})$ can be a natural approximation to $\delta(x)$ as the kernel width $\alpha \rightarrow 0$.
For $d$-dimensions space, $\eta_{\alpha}(\bx) = \alpha^{-d}\eta(\dfrac{\bx}{\alpha})$ can be used instead. 
In our experiments, we use Gaussian distribution, Cauchy distribution, or Laplacian distribution with sufficiently small kernel width $\alpha$ to approximate $\delta(\bx)$.

\subsection{Lower Bound Constrained Uncertainty Weighting}\label{subsec:MTL}
Our approximation to $\delta(x)$ eliminates function discontinuity at the origin, but it also brings a difficulty for the PINNs method to optimize its training loss. 
As the probability density function is highly concentrated at the origin as $\alpha \rightarrow 0$, the residual samples close to the origin tend to dominate the whole residual loss, making it difficult for the network to learn the governing equations and therefore the standard optimizer may not converge to the solution. 
To address this problem, we split the feasible region into two subdomains such that:
\[\Omega=\Omega_{0}\cup\Omega_{1}\quad \Omega_{0}\cap \Omega_{1} =\emptyset\]
where $\Omega_{0}$ represents the subdomain that contains the origin and $ \Omega_{1} $ covers the complement region. 
By taking this decomposition, the residual loss $L_{r}$ in Eq.\eqref{eq:PINN_loss} is divided into two parts:
\begin{equation}\label{eq:residual_loss_split}
L_r(\theta) = \lambda_{r, 0}L_{r,0}(\theta)  +\lambda_{r, 1}L_{r,1}(\theta) 
\end{equation}
where $L_{r,0}(\theta)$ and $L_{r,1}(\theta)$ correspond to the residual losses in $\Omega_{0}$ and $\Omega_{1}$, respectively. 
In our experiment, for the point source excited at $\bx_0 \in \Omega$, we split $\Omega$ as  
\[
\Omega_{0} = \{\bx_0+\bx\in \Omega\, | \,\|\bx\|\leq 3\alpha\},\quad \Omega_1 = \Omega \backslash\Omega_{0}.
\]
Together with the IC\&BC conditions, the total PINNs loss in Eq.\eqref{eq:PINN_loss_total} changes to be:
\begin{equation}\label{eq:new_PINN_loss}
	\begin{aligned}
	L_{total}(\theta) = \lambda_{r, 0}L_{r,0}(\theta)  +\lambda_{r, 1}L_{r,1}(\theta) + \lambda_{ic}L_{ic}(\theta)+\lambda_{bc}L_{bc}(\theta).
	\end{aligned}
\end{equation}
Properly setting the weight vector $\lambda = (\lambda_{r, 0}, \lambda_{r, 1}, \lambda_{ic}, \lambda_{bc})$ is critical to enhance the trainability of constrained neural networks \cite{mcclenny2020self, wang2020and},
but searching the optimal weight vector through manual tuning is infeasible. 
We adapt the uncertainty weighing method in \cite{kendall2018multi} to solve the problem.
For a total loss function including $m$ loss terms $L(\theta)=\sum_{i=1}^{m} \lambda_i L_i(\theta)$, we add a nonnegative $\epsilon^2$ to $w^2$, and get the total PINNs loss function as follows:
\begin{equation}\label{eq:lower_bounded_mtl}
	L_{total}(\theta; \bw) = \sum_{i=1}^{m} \dfrac{1}{2 (\epsilon^2 + w_i^2)} L_i(\theta) + \log(\epsilon^2+w_i^2)
\end{equation}
The uncertainty of the $i$-th loss term is approximated by $\epsilon^2 + w_i^2$, whose lower bound is constrainted by $\epsilon^2$ when decreasing $w_i$ to 0.
We argue that such reformulation is nontrivial since it provides a lower bound for the uncertainty estimation of each loss term and thus increases the stability of the self-adaptive weighting algorithm. 
In our experiment, taking $\epsilon=0.01$ leads to robust model outputs and outperforms the original formulation ($\epsilon=0$).

\begin{figure}[htbp]
	\centering
	\includegraphics[width=0.6\textwidth]{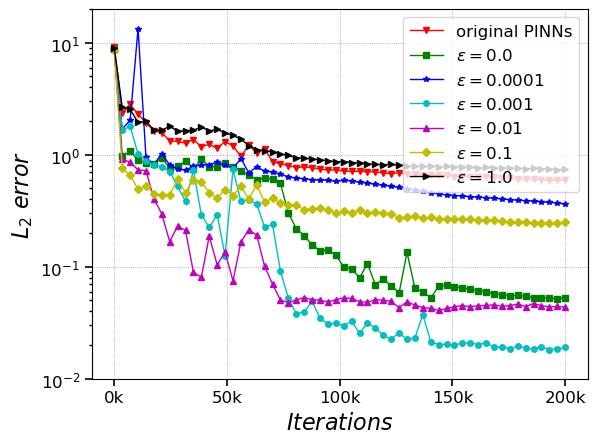}
	\caption{ \textbf{Maxwell's equations: }\normalsize $L_2\ error$s for different uncertainty lower bound.
		\textit{Original PINNs} means the equally weighted PINNs method.}
	\label{fig:ablation_mtl}
\end{figure}

Fig.\ref{fig:ablation_mtl} shows the $L_2\ error$s when different $\epsilon$ is used in solving Maxwell's equations with a point source (i.e., Eq.\eqref{def:maxwell_point_source}). 
For detailed experimental settings, please refer to Sec.\ref{subsec:maxwell_exp}.
When $\epsilon=0.01$ and $\epsilon=0.001$, the lower bound constrained uncertainty weighting method outperforms the original uncertainty weighting method ($\epsilon=0$) with not only lower final $L_2\ error$s, but also faster convergence speed.

\subsection{Multi-Scale DNN with Periodic Activations}\label{subsec:MscaleDNN}
When $\Omega=(0,\pi)^2$ and the point source is located at $(x_0,y_0)$, the analytical solution of the 2-D Poisson's equation with a point source (i.e., Eq.\eqref{def:possion_point_source}) is listed as follows: 
\begin{equation}
    u(x,y,x_0,y_0) = \frac{4}{\pi^2} \sum_{m,n=1}^{\infty}\frac{\sin{m x}\sin{m x_0}\sin{n y}\sin{n y_0}}{m^2 + n^2}
\end{equation}
The analytical solution is formed by superposition of multiple frequency components.
This multi-frequency phenomenon is common in solutions of many PDEs.
Recent development in deep learning theories \cite{rahaman2019spectral,xu2019training,xu2019frequency} reveals that the neural network fits the hidden physical law from low to high frequency components, which differentiates from the classical iterative numerical algorithms (e.g., Gauss-Seidel method).
Liu et al. \cite{liu2020multi} proposes multi-scale deep neural networks (MscaleDNNs) based on this theory which exhibit faster convergence in higher frequency components. 

Periodic nonlinearities have been investigated repeatedly over the past decades. 
Recently, the SIREN architecture \cite{sitzmann2020implicit} proposes to use \textit{Sine} function as a periodic activation function and outperforms alternative activation functions when dealing with implicit neural representations and their derivatives. 
Compared with other activation functions, e.g., \textit{Tanh, ReLU}, the derivative of a \textit{Sine} function is \textit{Cosine}, a phase-shifted \textit{Sine}. 
Therefore, the derivatives of a SIREN inherit the properties of SIREN, thus enabling us to supervise any derivative of SIREN with complex natural signals from PDEs.
However, the SIREN architecture may yield poor accuracy when applied to PDEs with a point source whose solution are formed by superposition of multiple frequency components. 

\begin{figure}[htbp]
	\centering
	\includegraphics[width=0.8\textwidth]{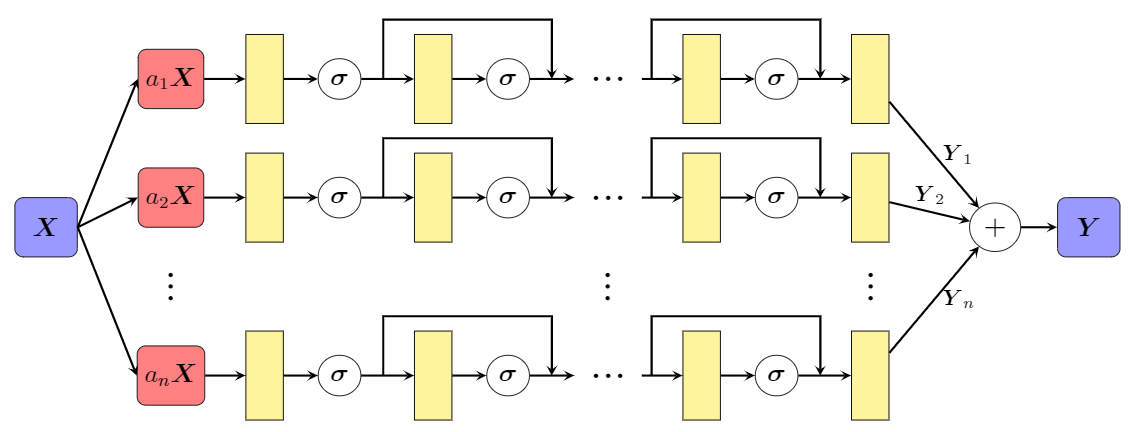}
	\caption{Architecture of MS-SIREN that consists of $n$ subnets with different scaling parameter $\{a_1, \ldots, a_n\}$ and the activation function $\sigma(\bx)=\sin(\bx)$.}
	\label{fig:multi_scale_siren}
\end{figure}

\begin{figure}[h]
	\centering
	\begin{subfigure}{0.48\textwidth}
		\centering
		\includegraphics[width=\textwidth]{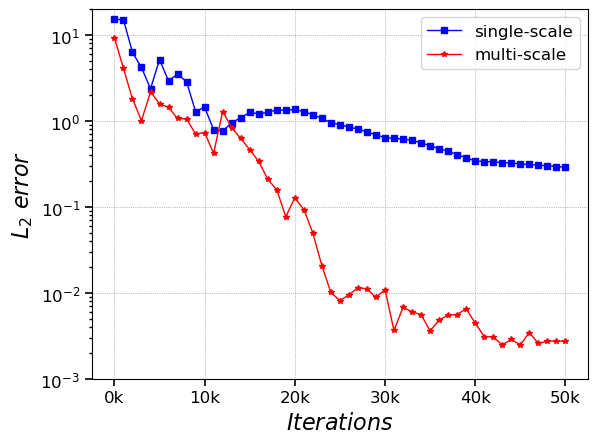}
		\caption{Single-scale VS. Multi-scale.}
		\label{fig:multi_scale_comparison_maxwell}
	\end{subfigure}
	\hfill
	\begin{subfigure}{0.48\textwidth}
		\centering
		\includegraphics[width=\textwidth]{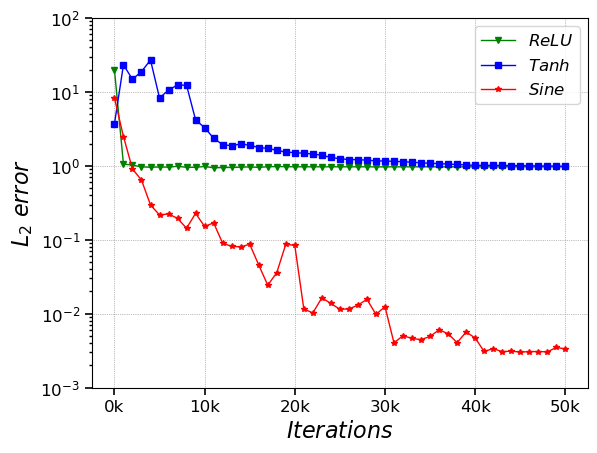}
		\caption{Activation functions.}
		\label{fig:activation_comparison_maxwell}
	\end{subfigure}
	\caption{\textbf{Poisson's equation: }\normalsize Comparative results under different architectures. (a) shows a single-scale network failing to converge to the solution. (b) shows the \textit{Sine} activation functions outperforming other alternatives.}
	\label{fig:two_graphs}
\end{figure}

Borrowing ideas from both SIREN and MscaleDNNs, we propose Multi-Scale SIREN (MS-SIREN) that combines a multi-scale DNN with \textit{Sine} activation function, as shown in Fig.\ref{fig:multi_scale_siren}. 
Skip connections are added between consecutive hidden layers to accelerate training and improve accuracy, which make the neural network much easier to train since they help to avoid the vanishing gradient problem \cite{he2016deep}.
Fig.\ref{fig:two_graphs} shows the effectiveness of MS-SIREN compared with other alternative architectures.

\section{Numerical Experiments}\label{sec:numerical_experiments} 
To evaluate the effectiveness of our proposed method, we apply it to solve the following three classical point source problems which come from different fields of physics: 
(1) electromagnetic simulation with the transverse electric (TE) mode \cite{gedney2011introduction};
(2) Poisson's equation with a point source;
(3) Barry and Mercer's source problem with time-dependent fluid injection \cite{bekele2020physics}. 
Unless otherwise specified,  we take the following default training settings:
\begin{itemize}
\item The Dirac delta function $\delta(\bx)$ is approximated by a Gaussian distribution with default kernel width $\alpha=0.01$.
\item $\epsilon=0.01$ is chosen for the lower bound constrained uncertainty weighting algorithm.
\item For MS-SIREN architecture, four subnets are used with scale coefficients set to \{1, 2, 4, 8\}. Each subnet contains seven fully connected layers and each layer contains 64 neurons. 
	  The exception is Poisson's equation, which requires only five fully connected layers per subnet to achieve ideal accuracy.
\end{itemize}
Average relative $L_2\ error$ is used to evaluate the performance of the trained network $u( \bx; \theta)$, which is computed as:
\begin{equation}
L_2\ error = \dfrac{\sum_{j=1}^{N}\|u_{ref}(\bx_j) - u(\bx_j, \theta)\|_2}{\sum_{j=1}^{N}\|u_{ref}(\bx_j)\|_2}
\end{equation}
where $N$ is the number of test points in $\Omega$ and $u_{ref}(\cdot)$ represents the ground truth. 
In our experiments, all feasible regions are rectangles, thus the spatial-temporal meshes are chosen as the test point set.
Unless otherwise specified, all the experiments are conducted under the MindSpore\footnote{https://www.mindspore.cn/} and trained on the Ascend 910 AI processors\footnote{https://e.huawei.com/en/products/servers/ascend}.
Readers can read our source code\footnote{https://gitee.com/mindspore/mindscience/tree/master/MindElec/} for implementation details.

\subsection{Electromagnetic Simulation with the TE Mode} \label{subsec:maxwell_exp}
Maxwell's equations are a set of coupled partial differential equations related to the fundamental physical modeling of electromagnetism. 
We use 2-D Maxwell's equations to validate the robustness of our method.
The governing equations of TE wave are given as:
\begin{subequations}
	\begin{align}
		\frac{\partial E_x}{\partial t } &= \frac{1}{\epsilon} \frac{\partial H_z}{\partial y}, \label{eq:maxwell_pde_1}\\
		\frac{\partial E_y}{\partial t } &= -\frac{1}{\epsilon} \frac{\partial H_z}{\partial x}, \label{eq:maxwell_pde_2}\\
		\frac{\partial H_z}{\partial t } &= -\frac{1}{\mu} (\frac{\partial E_y}{\partial x}- \frac{\partial E_x}{\partial y} + J). \label{eq:maxwell_pde_3}
	\end{align}
\end{subequations}
where $E_x$ and $E_y$ are the electric field in the Cartesian coordinate, and $H_z$ is the magnetic field perpendicular to the horizontal plane.
The constants $\mu \approx 4\pi \times 10^{-7} H/m$ and $\epsilon \approx 8.854 \times 10^{-12} F/m$ are known as the permeability and permittivity of the free space.
$J$ is a known source function that represents the energy passing through the source node. 
Without loss of generality, a Gaussian pulse containing multiple frequency components is used as the source function that is expressed as:
\begin{equation}
	\begin{aligned}
		J(x, y, t) = e^{-(\frac{t-d}{\tau })^2} \delta(x-x_0)\delta(y-y_0). \label{eq:gauss_pulse}
	\end{aligned}
\end{equation}
where $d$ is the temporal delay and $\tau$ is a pulse-width parameter. 
Currently, we take $\tau = 3.65 \times \sqrt{2.3}/(\pi f)$ with the characteristic frequency $f$ set to $1 GHz$.
Suppose the initial electromagnetic field is static, which corresponds to the initial conditions as follows:
\begin{subequations}
	\begin{align}
		E_x(x, y, 0) &= E_y(x, y, 0) = 0, \label{eq:maxwell_ic_1}\\
		H_z(x, y, 0) &= 0. \label{eq:maxwell_ic_2}
	\end{align}
\end{subequations}
To solve the equations uniquely over a finite vacuum domain, the standard Mur's second-order absorbing boundary condition is utilized. 
For a rectangular truncated domain, boundary conditions for the left, right, upper and lower sides, can be expressed as follows:
\begin{subequations}
	\begin{align}
		\frac{\partial H_z}{\partial x } - \frac{1}{c} \frac{\partial H_z}{\partial t} + \frac{c\epsilon}{2}\frac{\partial E_x}{\partial y} &= 0, \label{eq:maxwell_bc_1}\\
		\frac{\partial H_z}{\partial x } + \frac{1}{c} \frac{\partial H_z}{\partial t} - \frac{c\epsilon}{2}\frac{\partial E_x}{\partial y} &= 0, \label{eq:maxwell_bc_2}\\
		\frac{\partial H_z}{\partial y } - \frac{1}{c} \frac{\partial H_z}{\partial t} + \frac{c\epsilon}{2}\frac{\partial E_y}{\partial x} &= 0, \label{eq:maxwell_bc_3}\\
		\frac{\partial H_z}{\partial y } + \frac{1}{c} \frac{\partial H_z}{\partial t} - \frac{c\epsilon}{2}\frac{\partial E_y}{\partial x} &= 0. \label{eq:maxwell_bc_4}
	\end{align}
\end{subequations}
where $c = 1 / \sqrt{\mu \epsilon}$ is the light speed.

\begin{figure}[htbp]
	\centering
    \includegraphics[width=0.8\textwidth]{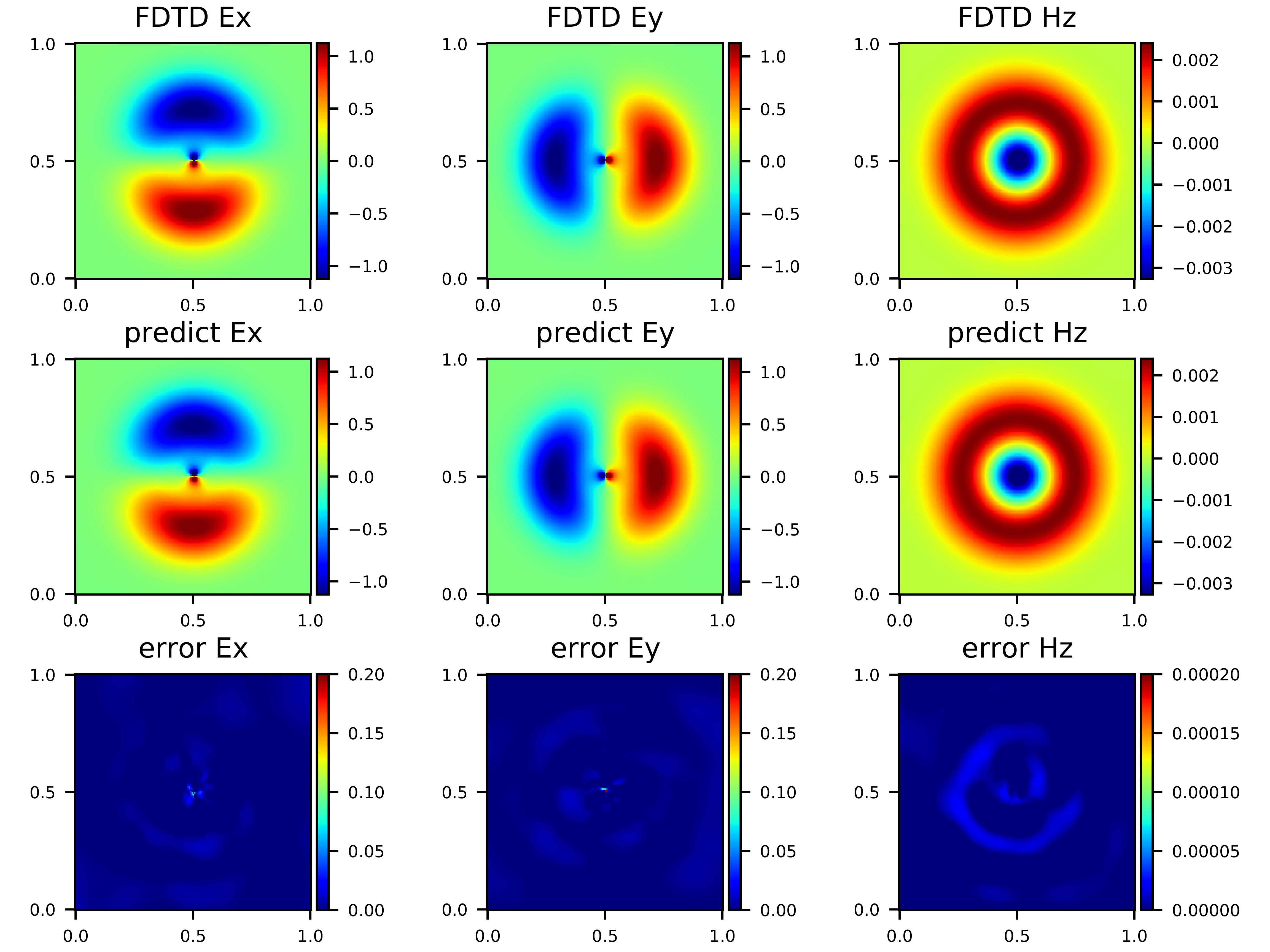}
	\caption{\normalsize \textbf{Maxwell's equations:} Solutions at the time of $2.4ns$.
		\textbf{Top:} The numerical solution ($E_x, E_y, H_z$) obtained by FDTD method;
		\textbf{Middle:} The predicted solution ($E_x, E_y, H_z$) output by our model;
		\textbf{Bottom:} The absolute error between model prediction and the reference solution.}
	\label{fig:instantaneous_field}
\end{figure}

We solve the problem in a rectangular domain $\Omega  = [-1, 1]^2$, and the point source is located at the origin $(x,y) = (0,0)$. 
The total simulation time is set to $4ns$ so that the pulse signals can propagate throughout the whole space of truncated domain. 
The reference solution is obtained through the finite-difference time-domain (FDTD) method \cite{gedney2011introduction} at a spatial resolution of $\Delta =0.005$. 
For model training, the spatial points are sampled from uniform distribution so that the average spacing between two adjacent random points is comparable with that in the reference FDTD mesh, and the average temporal step is chosen to satisfy the $CFL$ convergence condition \cite{gedney2011introduction}. 
The spatial point source is approximated by the Gaussian distribution with kernel width $\alpha=2\Delta=0.01$.
The model is trained using Adam optimizer with an initial learning rate of 0.001, and the learning rate attenuates 10 times when the training process reaches 40\%, 60\%, and 80\%, respectively.
The total number of iterations is $200$k, and the batch size is $N_{r,0}=N_{r,1}=N_{ic}=N_{bc}=8192$ for each iteration.
The instantaneous electromagnetic fields at the time of $2.4ns$ are showed in Fig.\ref{fig:instantaneous_field}. We can see that the predicted electromagnetic fields by our method are almost indistinguishable from the reference solution.

\noindent\textbf{Ablation Studies}

\begin{figure}[htbp]
	\centering
	\includegraphics[width=0.6\textwidth]{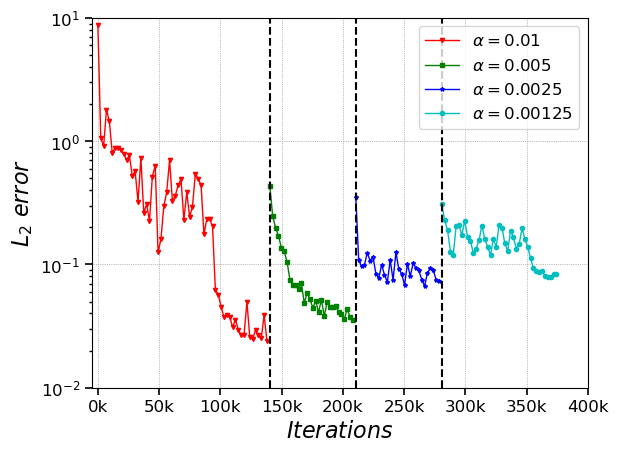}
	\caption{\textbf{Maxwell's equation: }\normalsize The mean $L_2\ error$ when different kernel widths are used. 
		The value of $\alpha$ is set to 0.01 for the first $140$k iterations and then halved every $70$k iterations.}
	\label{fig:convergence_of_variance}
\end{figure}

\noindent\textit{Selection of Kernel Width $\alpha$.}
In order to approximate the Dirac delta function, the setting of kernel width $\alpha$ is important.
In our experiments we find that the training process may not converge if $\alpha$ is too large or too small. 
We adopt an online fine-tuning method to decrease the value of $\alpha$ gradually during the training, which is illustrated in Fig.\ref{fig:convergence_of_variance} where the Gaussian distribution is used. 
The training process starts with $\alpha = 0.01$, the $L_2\ error$ drops rapidly and after $140$k iterations the final mean $L_2\ error$ is stable below 0.05. 
Then the online fine-tuning is performed to reduce the value of $\alpha$ by half, which makes the $L_2\ error$ rise first but then decrease quickly as the training progresses, and after $70$k iterations the final mean $L_2\ error$ is stable below 0.08. 
The fine-tuning operation repeats twice that reduces the value of $\alpha$ to 0.0025 and 0.00125, respectively. 
According to the experimental results in Fig.\ref{fig:convergence_of_variance}, we choose $\alpha = 0.01$ as the kernel width and fix it in the subsequent experiments. 
We further use Cauchy distribution and Laplacian distribution with $\alpha = 0.01$ to approximate the point source, and get the final mean $L_2\ error$s as 0.04 and 0.05, respectively. 
This set of experiments demonstrate that approximating the Dirac delta function with a symmetric unimodal continuous probability density function is feasible.

\noindent\textit{Network Architecture.}  
This set of experiments aim to validate the effectiveness of our proposed MS-SIREN architecture. 
The SIREN architecture (\# subnets = 1) and our MS-SIREN architecture are used to solve the Maxwell's equations with a point source under different settings, and the final $L_2\ error$s of all output components and their mean values are shown in Table \ref{tb:maxwell_ablation_mscale}. 
It is obvious that MS-SIREN outperforms SIREN under all settings, and the best performance is achieved by the architecture comprising four subnets.
\begin{table}[htbp]
	\centering
	\caption{\textbf{Maxwell's equations: } \normalsize $L_2\ error$ achieved by different network architectures.}
	\setlength{\tabcolsep}{1mm}{
		\begin{tabular}{c | c | c |c c c c}
			\hline
			\multicolumn{3}{c|}{network architectures} & \multicolumn{4}{c}{$L_2\ error$} \\
			\cline{1-7}
			$\#$ subnets & $\#$ layers per subnet & $\#$ neurons per layer&  $E_x$ & $E_y$ & $H_z$ & mean\\
			\hline
			1 & 7 & 256 & 0.192 & 0.183 & 0.433 & 0.269 \\
			1 & 9 & 256 & 0.147 & 0.146 & 0.070 & 0.121 \\
			\hline
			2 & 7 & 128 & 0.079 & 0.074 & 0.058 & 0.072 \\
			2 & 9 & 128 & 0.054 & 0.053 & 0.019 & 0.027 \\
			\hline
			4 & 7 & 64 & 0.021 & 0.022 & 0.001 & \textbf{0.018} \\
			4 & 9 & 64 & 0.025 & 0.022 & 0.017 & 0.021 \\
			\hline
	\end{tabular}}
	\label{tb:maxwell_ablation_mscale}
\end{table}

\begin{figure}[htbp]
	\centering
	\includegraphics[width=0.6\textwidth]{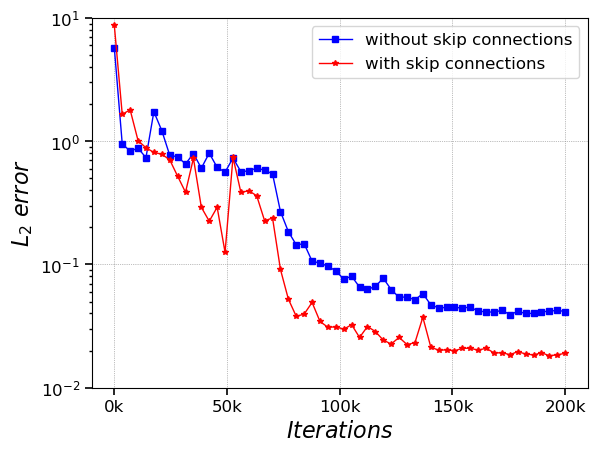}
	\caption{\textbf{Maxwell's equations: } \normalsize The mean $L_2\ error$s got by MS-SIREN with/without skip connections.}
	\label{fig:convergence_of_residual}
\end{figure}

The MS-SIREN architecture uses skip connections between consecutive hidden layers.
To validate its effectiveness, we compare MS-SIREN architectures with and without skip connections, and show the results in Fig.\ref{fig:convergence_of_residual}. 
It is clear that skip connections make $L_2\ error$ decrease faster and reach a lower value.

\begin{figure}[htbp]
	\centering
	\includegraphics[width=0.6\textwidth]{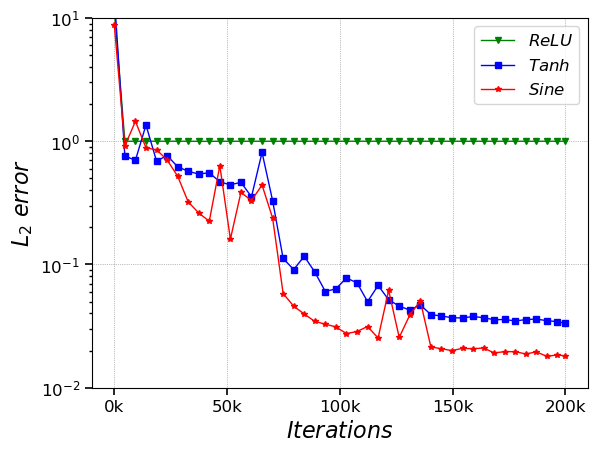}
	\caption{\textbf{Maxwell's equations: } \normalsize Influence of activation functions on the prediction accuracy.}
	\label{fig:convergence_of_activations}
\end{figure}

Activation function is another key factor affecting the prediction accuracy. 
To validate the effectiveness of \textit{Sine} periodic activation function, we compare it with \textit{ReLU} and \textit{Tanh}, and the results are shown in Fig.\ref{fig:convergence_of_activations}. 
\textit{Sine} produces the best results in terms of convergence speed and the final mean $L_2\ error$. 
In comparison, \textit{Tanh} performs slightly worse than \textit{Sine}, and \textit{ReLU} fails to let the network learn any useful information.

\subsection{Poisson's Equation with a point source}
\begin{figure}[htbp]
	\centering
	\includegraphics[width=0.8\textwidth]{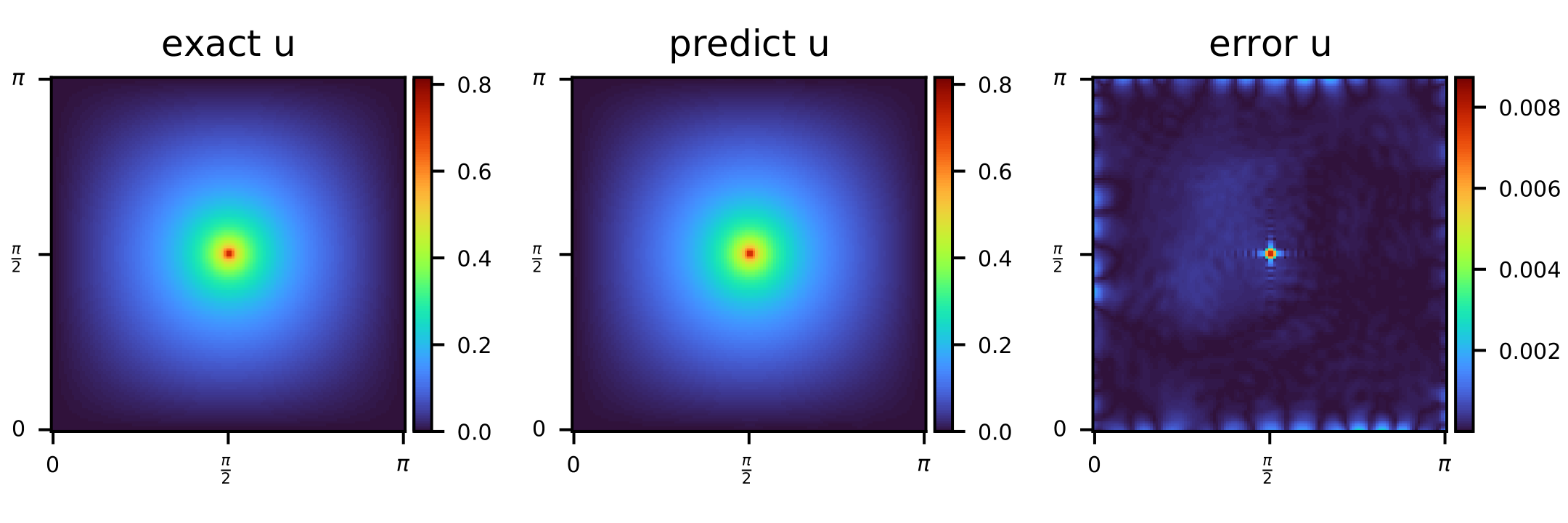}
	\caption{\textbf{Poisson's Equation: }\normalsize Analytical solution, model prediction and absolute error (from left to right).}
	\label{fig:poisson_compare_exact}
\end{figure}

We apply our approach to solve Eq.\eqref{def:possion_point_source} in the region of $\Omega=[0,\pi]\times [0,\pi]$ with the point source located at $\bx_0=(\frac{\pi}{2}, \frac{\pi}{2})$.
The point source is approximated using Gaussian distribution with fixed $\alpha = 0.01$ and the lower bound hyperparameter $\epsilon$ is set to 0.01.
The model is trained for $50k$ iterations using Adam optimizer with an initial learning rate of 0.001, and the learning rate attenuates 10 times when the training process reaches 40\%, 60\%, and 80\%, respectively.
In each iteration, the batch size is set to $(N_{r_,0}, N_{r,1}, N_{bc})=(8192, 8192, 8192)$. 
Fig.\ref{fig:poisson_compare_exact} shows that the absolute error between our solution and the analytical solution is small.

\begin{figure}[h]
	\centering
	\begin{subfigure}{0.58\textwidth}
		\centering
		\includegraphics[width=\textwidth]{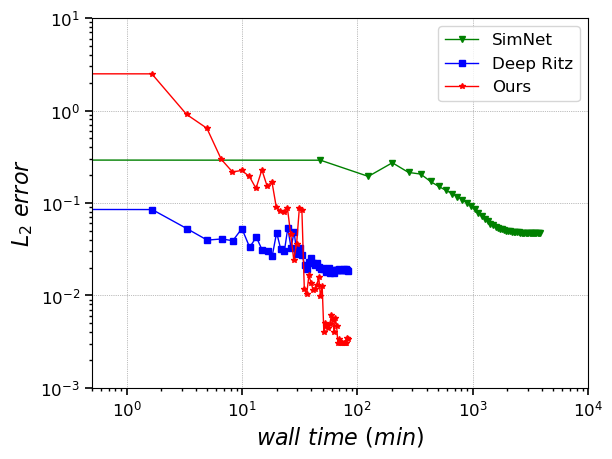}
		\caption{The $L_2\ error$s convergence with wall time.}
		\label{fig:poisson_ablation_other_method}
	\end{subfigure}
	\hfill
	\begin{subfigure}{0.38\textwidth}
		\centering
		\includegraphics[width=\textwidth]{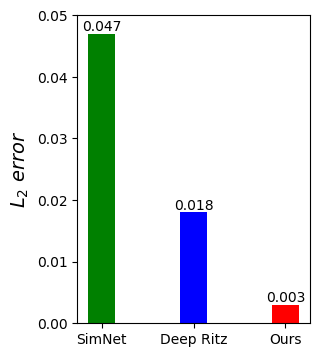}
		\caption{The final $L_2\ error$s.}
		\label{fig:poisson_ablation_other_method_box}
	\end{subfigure}
	\caption{\textbf{Poisson's Equation: } \normalsize Comparison of different methods.
	}
	\label{fig:poisson_ablation_other_method_two}
\end{figure}

For comparisons, we also apply the Deep Ritz method and the variational method in SimNet to solve the same problem.  
In this set of experiments, we implement our method and Deep Ritz method in MindSpore framework, and implement the variational method using the public code in SimNet with a default selection of 15 test functions.
All the experiments are run on an NVIDIA Tesla V100 GPU card.
Fig.\ref{fig:poisson_ablation_other_method} shows that the convergence speed of our method is close to that of Deep Ritz method, but faster than that of SimNet.
Fig.\ref{fig:poisson_ablation_other_method_box} shows the final $L_2\ error$s obtained by different methods, and obviously our method achieves the highest accuracy.

\noindent\textbf{Ablation Studies of Network Architectures.} 
Table \ref{tb:poisson_ablation_mscale} shows the prediction accuracy obtained by different network architectures. 
The SIREN architecture (\# subnets = 1) cannot converge in all settings due to the singularity problem. 
However, when the number of subnets increases to 2 or 4, the $L_2\ error$ drops remarkably. 
This experiment indicates that a multi-scale structure is necessary for solving with this problem.

\begin{table}[htbp]
	\centering
	\caption{\textbf{Poisson's Equation: }\normalsize $L_2\ error$s obtained by different network architectures.}
	\begin{tabular}{c | c | c |c}
		\hline
		\multicolumn{3}{c|}{Network Architectures} & \multirow{2}{*}{$L_2\ error$}\\
		\cline{1-3}
		\# subnets & \# layers per subent & \# neurons per layer& \\
		\hline
		1 & 5 & 256 & 0.289\\
		1 & 7 & 256 & 1.081\\
		\hline
		2 & 5 & 128 & 0.003\\
		2 & 7 & 128 & 0.006\\
		\hline
		4 & 5 & 64 & 0.003\\
		4 & 7 & 64 & \textbf{0.002}\\
		\hline
	\end{tabular}
	\label{tb:poisson_ablation_mscale}
\end{table}


\subsection{Barry and Mercer's Source Problem}
The third problem we solve is the nondimensionalized poroelastic Barry and Mercer's source problem with time-dependent fluid injection.
The governing equations are given by
\begin{subequations}\label{eq:barry_mercer}
	\begin{align}
		\frac{\partial ^ 2 u}{\partial t \partial x} + \frac{\partial ^ 2 v}{\partial t \partial z} - \frac{\partial ^ 2 p}{\partial x^2} - \frac{\partial ^ 2 p}{\partial z^2} - \beta Q  & = 0,\label{eq:barry_mercer_pde_1}\\
		(\eta + 1)\frac{\partial ^ 2 u}{\partial x^2} + \frac{\partial ^ 2 u}{\partial z^2} + \eta \frac{\partial ^ 2 v}{\partial x \partial z} - (\eta + 1) \frac{\partial p}{\partial x} & = 0,\label{eq:barry_mercer_pde_2}\\
		\frac{\partial ^ 2 v}{\partial x^2} + (\eta + 1)\frac{\partial ^ 2 v}{\partial z^2} + \eta \frac{\partial ^ 2 u}{\partial x \partial z} - (\eta + 1) \frac{\partial p}{\partial z} & = 0.\label{eq:barry_mercer_pde_3}
	\end{align}
\end{subequations}
where $u$ and $v$ are the deformations of porous medium along the x and z directions, $p$ is the pore fluid pressure, $Q$ is a fluid source or sink term, $\eta$ and $\beta$ are nondimensional parameters which are functions of the real material parameters.

\begin{figure}[htbp]
	\centering
	\includegraphics[width=0.5\textwidth]{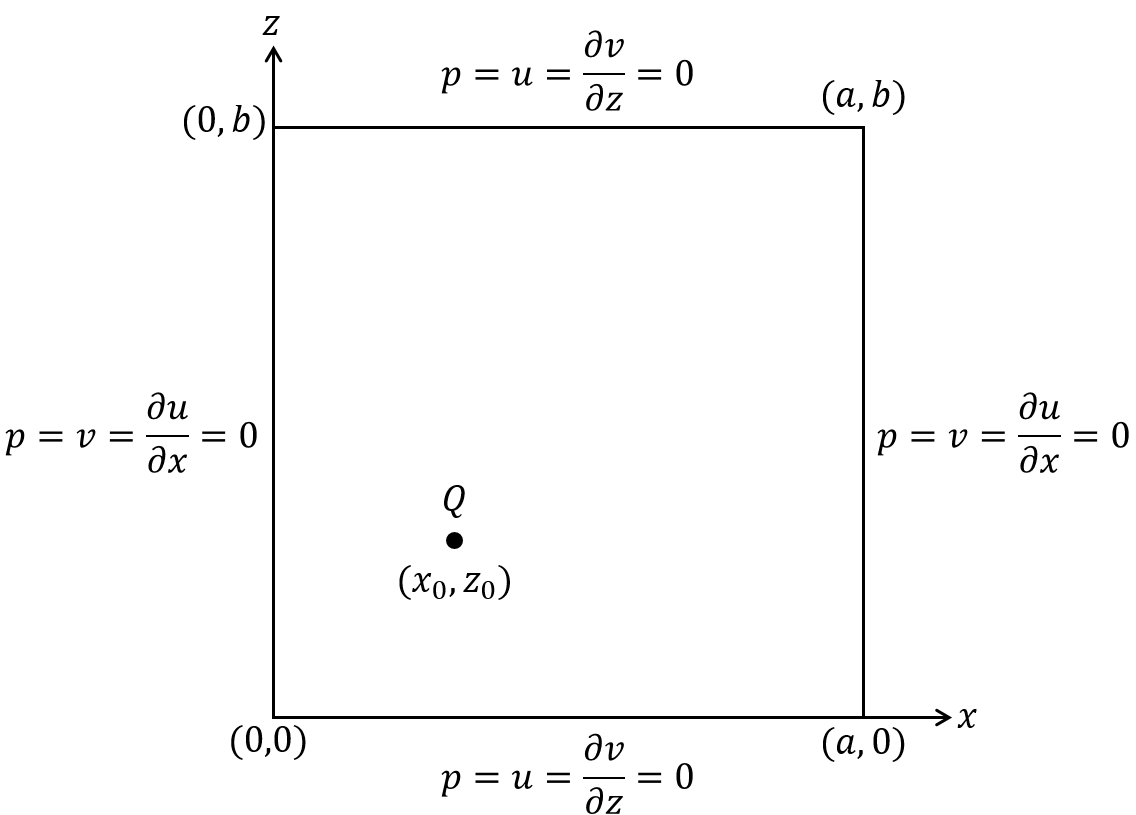}
	\caption{\textbf{Barry and Mercer’s source problem: }\normalsize Illustration of the computational domain and boundary conditions.}
	\label{fig:barry_mercer_problem_setting}
\end{figure}
An illustration of the problem is shown in Fig.\ref{fig:barry_mercer_problem_setting}. 
An oscillating point source $Q$ located at $(x_0, z_0)$ in the rectangular domain $[0, a]\times[0, b]$ is given by:
\begin{equation}\label{eq:barry_mercer_src}
	Q(x,z,t) = \delta(x-x_0)\delta(z-z_0)sin(\omega t)
\end{equation}
where $\omega$ is the oscillation frequency.
The parameters are set as follows: $a=b=1$, $t\in[0,2\pi]$, $\beta=2$, $\eta=1.5$, $(x_0,z_0)=(0.25,0.25)$ and $\omega=1$. 
The model is trained for $250k$ iterations using Adam optimizer with an initial learning rate of 0.001, and the learning rate attenuates 10 times when the training process reaches 40\%, 60\%, and 80\%, respectively.
In each iteration, the batch size is set to $(N_{r_,0}, N_{r,1}, N_{bc}, N_{ic})=(8192, 8192, 8192, 8192)$.

\begin{figure}[htbp]
	\centering
	\includegraphics[width=0.8\textwidth]{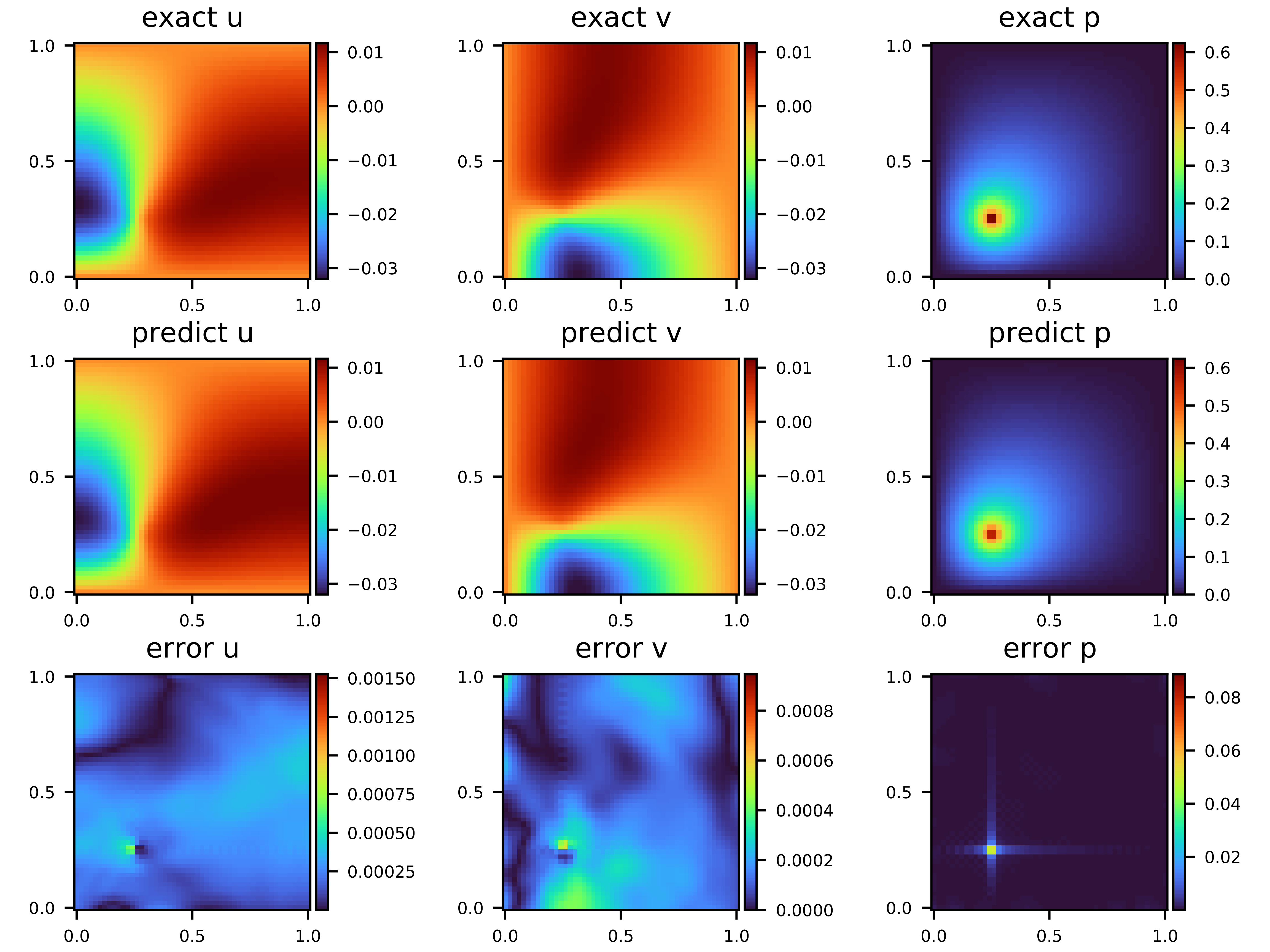}
	\caption{\normalsize \textbf{ Barry and Mercer's source problem:} Solutions at $t=\pi/5$.
		\textbf{Top:} Analytical solution;  \textbf{Middle:} Model prediction; \textbf{Bottom:} Absolute error.}
	\label{fig:barry_mercer_compare_exact}
\end{figure}

Fig.\ref{fig:barry_mercer_compare_exact} shows the solution at $t=\pi/5$ obtained by our method (middle) and the analytical solution (top), and the absolute error between them (bottom). 
The time averaged mean $L_2\ error$ of the predicted solution is 0.017, demonstrating the high accuracy of our method. 
In addition to Gaussian distribution, we also use Cauchy distribution and Laplacian distribution to approximate the Dirac delta function, and obtain similar accuracy with $L_2\ error$ being 0.051 and 0.025, respectively.

\noindent\textbf{Ablation Studies} 

\begin{figure}[htbp]
	\centering
	\includegraphics[width=0.6\textwidth]{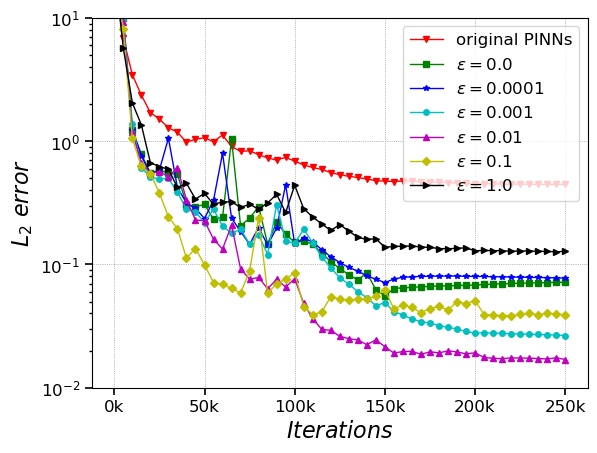}
	\caption{\textbf{Barry and Mercer’s source problem: }\normalsize Convergence speed of the mean $L_2\ error$s with different lower bound hyperparameter $\epsilon$.}
	\label{fig:barry_mercer_MTL}
\end{figure}

\noindent\textit{Lower Bound Hyperparameter $\epsilon$.}
Fig.\ref{fig:barry_mercer_MTL} shows the convergence speed of the mean $L_2\ error$s with different lower bound hyperparameter $\epsilon$, and the best performance (mean $L_2\ error = 0.017$) is achieved when $\epsilon=0.01$. 
It shows that our lower bound constrained uncertainty weighting method outperforms the uncertainty weighting method ($\epsilon=0$) \cite{kendall2018multi} and the equally weighted PINNs method (’original PINNs’).

\noindent\textit{Network Architectures.}
Table \ref{tb:barry_mercer_ablation_mscale} shows the $L_2\ error$s achieved by different network architectures.
Compared with the SIREN architecture (\# subnets = 1), MS-SIREN does not bring any benefit in this case as the point source in this problem contains only one frequency component, and thus there are not multiple frequency components in the analytical solution.
\begin{table}[htbp]
 	\centering
 	\caption{\textbf{Barry and Mercer’s source problem: } \normalsize $L_2\ error$s got by different network architectures.}
 	\setlength{\tabcolsep}{1mm}{
 		\begin{tabular}{c | c | c | c c c c}
 			\hline
 			\multicolumn{3}{c|}{Network Architectures} & \multicolumn{4}{c}{$L_2\ error$}\\
 			\cline{1-7}
 			\# subnets & \# layers per subnet & \# neurons per layer & $u$ & $v$ & $p$ & mean\\
 			\hline
 			1 & 5 & 256 & 0.022 & 0.021 & 0.026 & 0.023 \\
 			1 & 7 & 256 & 0.011 & 0.016 & 0.026 & 0.018 \\
 			\hline
 			2 & 5 & 128 & 0.010 & 0.011 & 0.026 & \textbf{0.016} \\
 			2 & 7 & 128 & 0.012 & 0.013 & 0.026 & 0.017 \\
 			\hline
 			4 & 5 & 64 & 0.012 & 0.013 & 0.026 & 0.017 \\
 			4 & 7 & 64 & 0.018 & 0.014 & 0.026 & 0.019 \\
 			\hline
 	\end{tabular}}
 	\label{tb:barry_mercer_ablation_mscale}
 \end{table} 

\begin{figure}[htbp]
	\centering
	\includegraphics[width=0.6\textwidth]{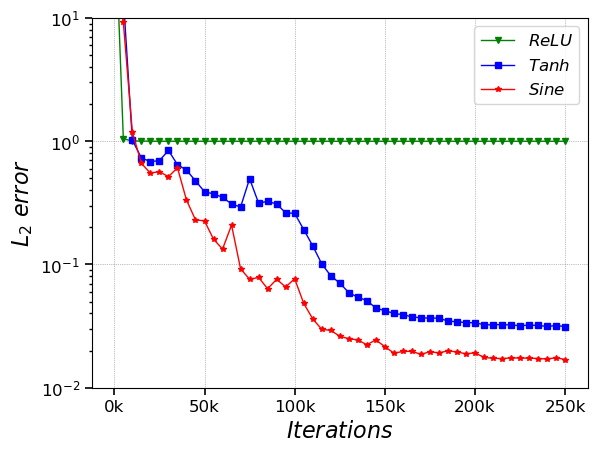}
	\caption{\textbf{Barry and Mercer’s source problem: } \normalsize Convergence speed of the mean $L_2\ error$s when different activation functions are used.}
	\label{fig:barry_mercer_ablation_act}
\end{figure}

Fig.\ref{fig:barry_mercer_ablation_act} compares the convergence speed of the mean $L_2\ error$s when different activation functions are used.  
The periodic activation function \textit{Sine} exhibits superior performance over \textit{ReLU} and \textit{Tanh} similar to the case of Maxwell’s equations.

In summary, approximating the Dirac function with the symmetric unimodal continuous probability density function is necessary for all experiments in order for PINNs to solve the singularity problem posed by the point source.
For the effectiveness of the MS-SIREN architecture, the multi-scale network exhibits better accuracy and faster convergence speed than the single-scale network in both Maxwell's equations and Poisson's equation experiments. 
However, the muti-scale network does not bring significant gain in Barry and Mercer's source problem due to the fact that the PDE solutions do not contain multiple frequency components.
Moreover, the \textit{Sine} activation function exhibits better accuracy than \textit{ReLU} and \textit{Tanh} in all three sets of experiments.
For the Maxwell's equations and the Barry and Mercer's source problem, the lower bound constrained uncertainty weighting algorithm can obtain a lower $L_2\ error$ relative to the original uncertainty weighing method.
It is worth noting that all three of our proposed techniques need to be utilized to obtain the desired accuracy when solving the Maxwell's equations.
\section{Conclusions} \label{sec:conclusions}

PDEs with a point source pose a great challenge to conventional PINNs method due to the singularity problem, and all the methods trying to tackle this problem so far can only work in some specific situations. 
We propose a universal approach based on the PINNs method that can solve the PDEs with a point source without any specific assumptions. 
The novelty of our approach lies in three techniques, approximating the Dirac delta function with a symmetric unimodal continuous probability density function, balancing the training loss term of different areas with a lower bound constrained uncertainty weighting algorithm, and using multi-scale DNN with \textit{Sine} activation function to build the network architecture.
Three representative physical problems are used to verify the effectiveness of our method, and these experiments show that our method outperforms existing deep learning-based methods in accuracy, efficiency and versatility. 
In the future we will try to extend the idea of smoothing function to solve discontinuous PDEs.
In addition, the lower bound constrained uncertainty weighting algorithm is not only restricted to solving PDEs, but also can be applied to many other multitask learning problems  in computer vision and natural language processing \cite{kendall2018multi}. 
An independent study on this direction is left for future explorations.

\bibliographystyle{unsrt}
\bibliography{ref}

\section{Appendix}

\vspace{0.1in}
\noindent{\textbf{Analytical Solution of Barry and Mercer's Source Problem}}\newline



For the Barry and Mercer’s Source Problem defined by Eq.\eqref{eq:barry_mercer} with the domain and boundary conditions shown in Fig.\ref{fig:barry_mercer_problem_setting}, the analytical solution is expressed as:

\begin{subequations}
\begin{align}
u(x,z,t) &= \frac{4}{ab}\sum_{n,q=1}^{\infty}\hat{u}(n,q,t)\cos{\lambda_n x}\sin{\lambda_q z}, \\
v(x,z,t) &= \frac{4}{ab}\sum_{n,q=1}^{\infty}\hat{v}(n,q,t)\sin{\lambda_n x}\cos{\lambda_q z}, \\
p(x,z,t) &= - \frac{4}{ab}\sum_{n,q=1}^{\infty}\hat{p}(n,q,t)\sin{\lambda_n x}\sin{\lambda_q z}.
\end{align}
\end{subequations}
where $\hat{u}$, $\hat{v}$, $\hat{p}$ are intermediate variables in the solution and can be expressed as follows:
\begin{subequations}
\begin{align}
\hat{u}(n,q,t) &= \frac{\lambda_n}{\lambda_{n q}} \hat{p}(n,q,t), \\
\hat{v}(n,q,t) &= \frac{\lambda_q}{\lambda_{n q}} \hat{p}(n,q,t), \\
\hat{p}(n,q,t) &= - \frac{\beta \sin{\lambda_n x_0} \sin{\lambda_q z_0}}{\lambda_{n q}^2} (\lambda_{n q} \sin{\omega t} - \omega \cos{\omega t} + \omega e^{-\lambda_{n q} t})
\end{align}
\end{subequations}
where
\begin{equation}
\begin{aligned}
\lambda_n = \frac{n \pi}{a},\quad \lambda_q = \frac{q \pi}{b},\quad \lambda_{n q} = \lambda_n^2 + \lambda_q^2.
\end{aligned}
\end{equation}

\end{document}